\pdfoutput=1

\documentclass[11pt]{article}
\usepackage{amssymb}
\usepackage{subfig}
\usepackage[final]{acl}
\usepackage{xcolor}

\usepackage{times}
\usepackage{latexsym}

\usepackage[T1]{fontenc}

\usepackage[utf8]{inputenc}

\usepackage{microtype}

\usepackage{inconsolata}

\usepackage{graphicx}

\usepackage{colortbl}
\usepackage{times}
\usepackage{latexsym}
\usepackage{tabularx}
\usepackage{booktabs}
\usepackage{amsmath}
\usepackage{float}
\usepackage{multirow}
 
\title{MEDA: Dynamic  KV Cache   Allocation for  Efficient \\  Multimodal Long-Context Inference}

\author{
    Zhongwei Wan\textsuperscript{1}, Hui Shen\textsuperscript{1}, Xin Wang\textsuperscript{1}, Che Liu\textsuperscript{2}, Zheda Mai\textsuperscript{1}, Mi Zhang\textsuperscript{1} \\
    \textsuperscript{1}The Ohio State University \quad
    \textsuperscript{2}Imperial College London \\
    \url{https://github.com/AIoT-MLSys-Lab/MEDA}
}

\begin{document}
\maketitle
\begin{abstract}

Long-context Multimodal Large Language Models (MLLMs) that incorporate long text-image and text-video modalities, demand substantial resources as their multimodal Key-Value (KV) caches grow with increasing input lengths, challenging inference efficiency.
%
Existing methods for KV cache compression, in both text-only and multimodal LLMs, have neglected attention density variations across layers, thus often adopting uniform or progressive reduction strategies for layer-wise cache allocation. 
%
In this work, we propose \textbf{MEDA}, a 
dynamic layer-wise KV cache allocation method for efficient multimodal long-context inference.
%
As its core, MEDA utilizes cross-modal attention entropy to determine the KV cache size at each MLLMs layer.
%
Given the dynamically allocated KV cache size at each layer, MEDA also employs a KV pair selection scheme to identify which KV pairs to select and a KV pair merging strategy that merges the selected and non-selected ones to preserve information from the entire context.
%
MEDA achieves up to \textbf{72\%} KV cache memory reduction and \textbf{2.82} times faster decoding speed, while maintaining or enhancing performance on various multimodal tasks in long-context settings, including multi-images and long-video scenarios. Our code is released at \url{https://github.com/AIoT-MLSys-Lab/MEDA}. 
\end{abstract}

\section{Introduction}
\label{sec: intro}
\begin{figure}[t]
    \centering
    \includegraphics[width=0.475\textwidth]{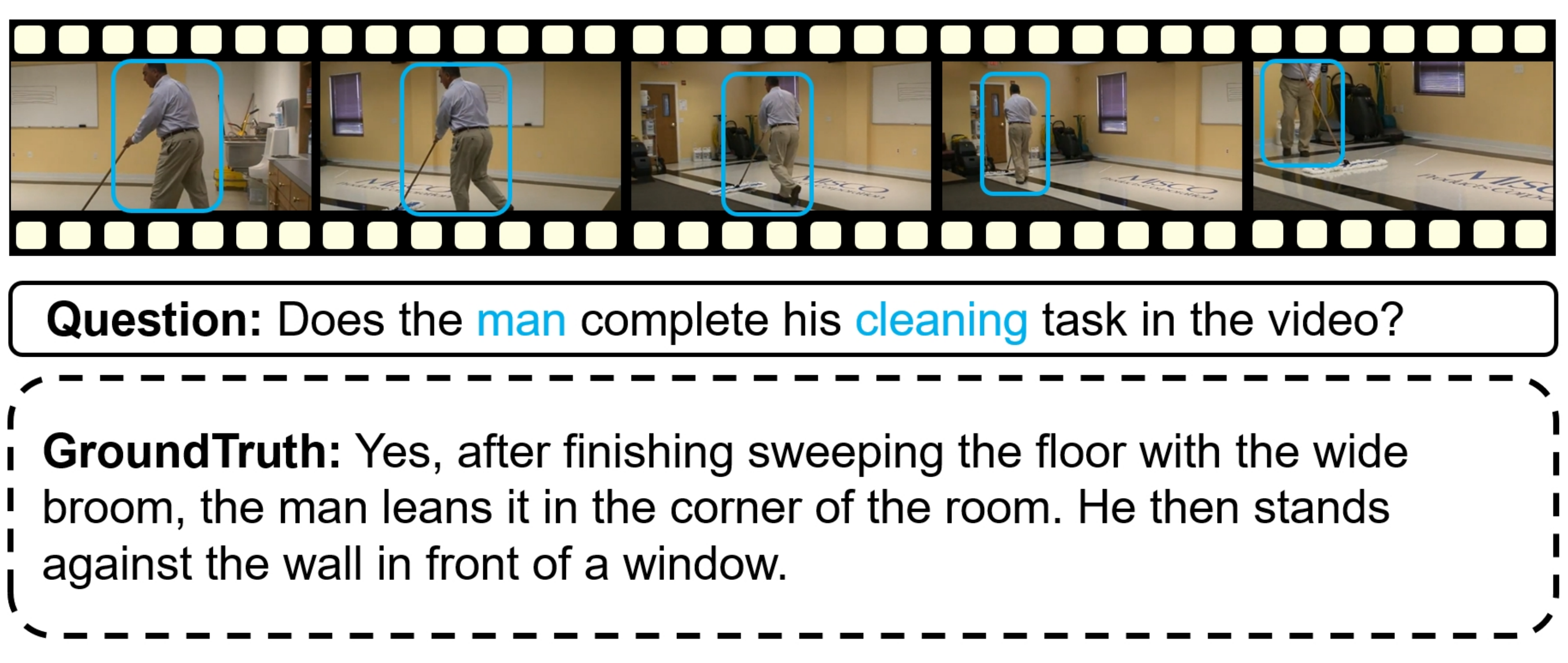}
    \vspace{-1mm}
    \caption{\small{A multimodal long-context sample from Video-ChatGPT~\cite{maaz2023video}, showing key information interactions between blue-boxed video frames and textual phrases.}}
    \vspace{-1mm}
    \label{fig: sample1}
\end{figure}

Long-context Multimodal Large Language Models (MLLMs) have achieved remarkable progress in processing multimodal long context involving long text-image and text-video inputs, as exemplified by LLaVA-NeXT~\cite{liu2024llava}, GPT-4V~\citep{achiam2023gpt} and long-video MLLMs~\cite{xue2024longvila, zhang2024long}. These models are capable of handling complex multimodal patterns within their Key-Value (KV) caches, such as text accompanied by multiple interrelated images or lengthy video sequences, introducing intricate cross-modal interactions, as shown in Figure~\ref{fig: sample1}. 
However, despite these advancements,
long-context MLLMs demand substantial resources as their multimodal KV caches grow with increasing input lengths in long-context settings, causing significant slowdown during inference.

%

\begin{figure}[t]
    \centering
    \subfloat{
        \includegraphics[width=0.485\textwidth]{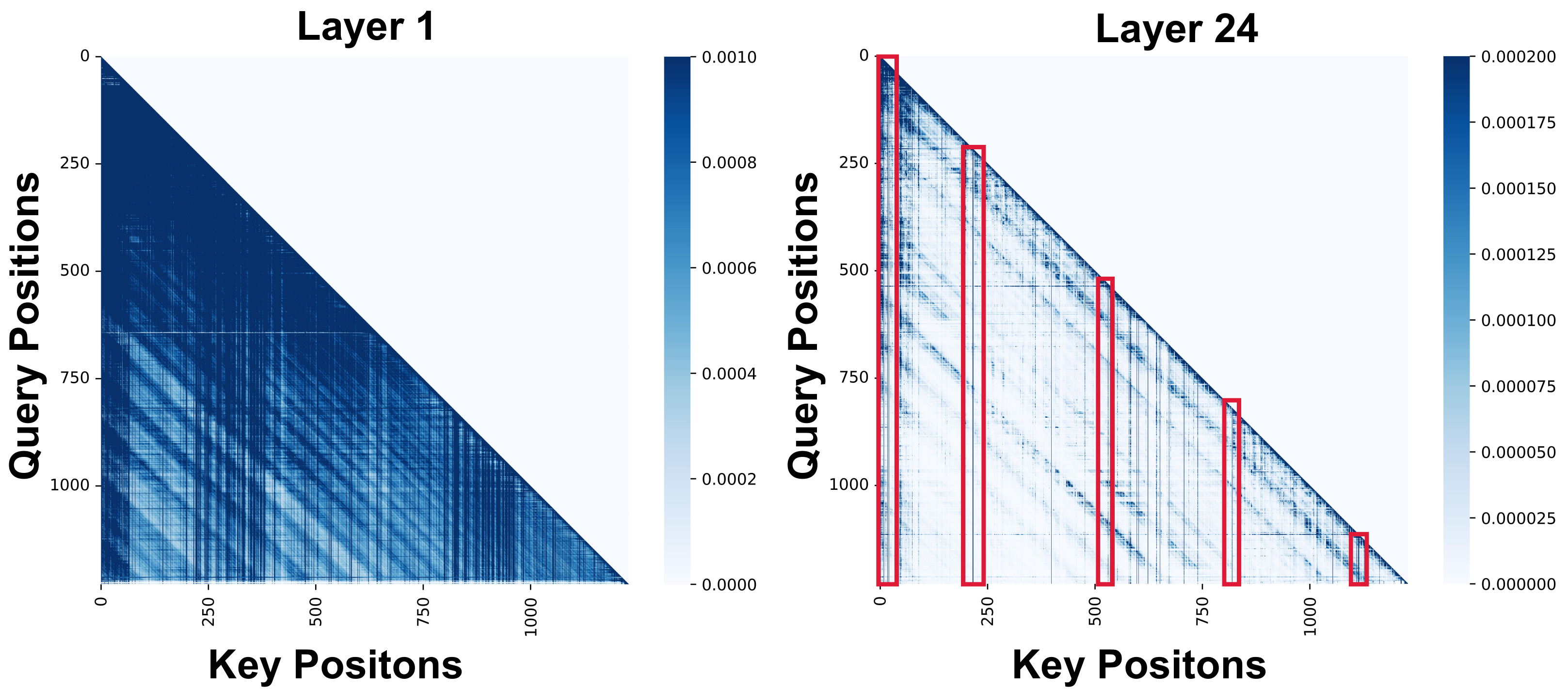}
        \vspace{-0.2in} 
    } \\ 
    \subfloat{
        \includegraphics[width=0.485\textwidth]{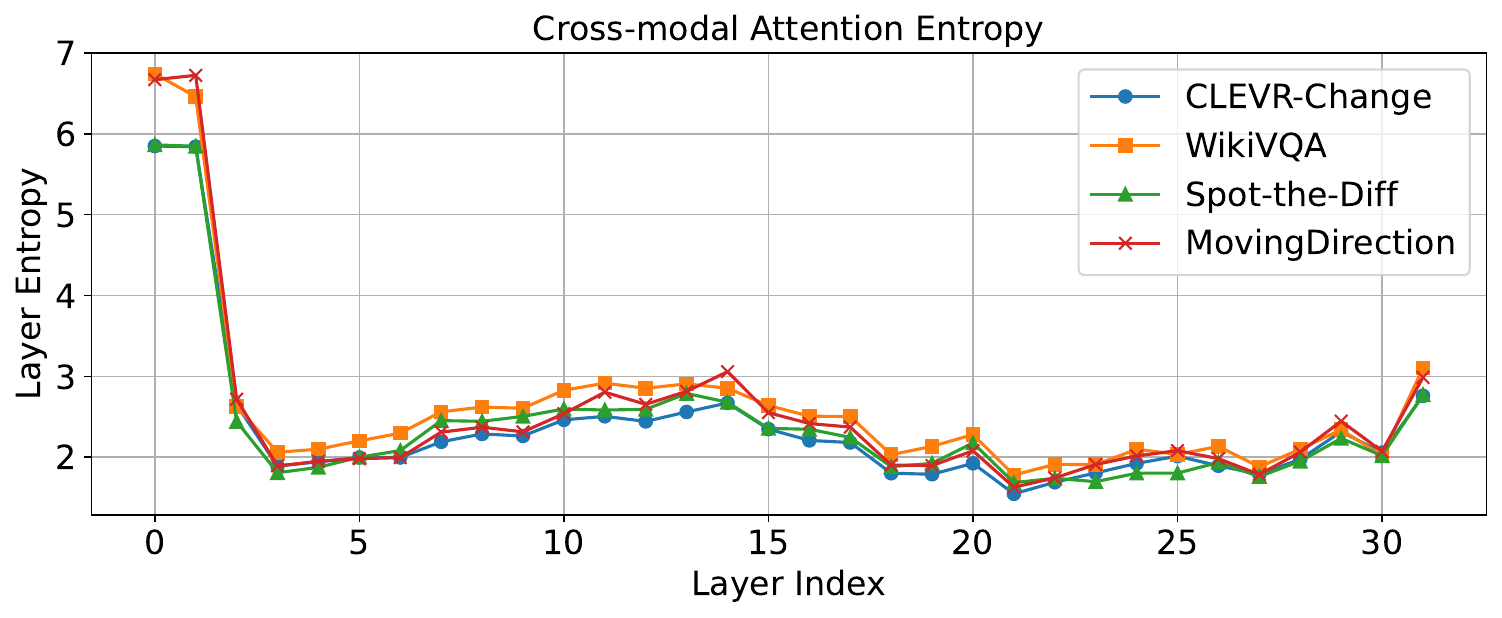}
        \vspace{-0.3in}
    }
    \caption{\small{Using the cross-modal attention entropy from Eq.~\ref{eq: Entropy}, we analyze LLaVA-NeXT-7B~\cite{liu2024llava} across different sub-tasks~\cite{Song2024MileBenchBM}. We observe varying multimodal interaction patterns: early layers (e.g., Layer 1) exhibit dense attention weights with higher entropy, while deeper layers (e.g., Layer 24) exhibit sparse attention weights with lower entropy, given that they focus on key tokens (red columns), similar to the blue areas and text in Figure~\ref{fig: sample1}.
   }}
    \vspace{-2mm}
    \label{fig: sample_combined}
\end{figure}

Conventional KV cache methods designed for text-only LLMs are difficult to be directly adopted to long-context multimodal inputs because they do not account for the complex cross-modal interactions present in long-context settings. Previous methods for KV cache compression in both text-only LLMs and MLLMs such as text-centric eviction-based methods~\citep{Zhang2023H2OHO, Ren2024OnTE, Li2024SnapKVLK}, static progressive layer-wise reduction methods~\cite{zhang2024pyramidkv, yang2024pyramidinfer}, and multimodal pruning methods~\cite{wan2024look} have predominantly employed \textit{uniform} cache size allocation across layers. 
%
However, these methods overlook the variations in attention density across different layers as illustrated in Figure~\ref{fig: sample_combined}.
As a consequence, allocating a uniform KV cache size across different layers, without accounting for these variations, can not only lead to information loss in dense layers if less KV caches are allocated, resulting in reduced precision and suboptimal performance, but also cause significant inefficiency in sparse layers when more than enough KV caches are allocated.

In this paper, we propose a \textit{dynamic} layer-wise KV cache allocation method, which we refer to as \textbf{MEDA} (\textbf{M}ultimodal Attention \textbf{E}ntropy-Guided \textbf{D}ynamic KV Cache \textbf{A}llocation), for efficient multimodal long-context inference.
%
The key idea of MEDA is that it proposes to use cross-modal attention entropy to capture the diverse cross-modal attention patterns at different layers in MLLMs, and then dynamically allocates KV caches across layers so as to adapt to the unique layer-wise attention distributions. 
%
Moreover, given the dynamically allocated KV cache size at each layer, MEDA employs a multimodal KV pair selection scheme to identify which KV pairs to be selected at each layer. For each KV pair that is not selected, MEDA incorporates a KV pair merging strategy that merges the selected and non-selected KV pairs to preserve information from the entire context despite the reduced KV cache size.
In doing so, MEDA is able to achieve efficient KV cache usage for multimodal long-context inference.
%
It is also worthwhile to note that MEDA does not require additional fine-tuning and can be seamlessly integrated as a plug-and-play solution, offering a dynamic KV cache allocation strategy tailored for multimodal contexts. 

We evaluate MEDA across various recent MLLM backbones, including LLaVA-v1.5-13B~\cite{Liu2023VisualIT}, LLaVA-NeXT-7B~\cite{liu2024llava}, and InternVL-v1.5-7B~\cite{Chen2023InternVLSU} for multi-images tasks, as well as LLaVA-Video-7B/32B~\cite{zhang2024video}, LongVA-7B~\cite{zhang2024long}, and LongVILA-8B~\cite{xue2024longvila} for long-video tasks. 
We also evaluate MEDA on diverse mutlimodal long-context datasets including MileBench~\citep{Song2024MileBenchBM}, Video-ChatGPT~\cite{maaz2023video}, DREAM-1K~\cite{wang2024tarsier}, and WorldQA~\cite{zhang2024worldqa}.
%
Our results show that MEDA outperforms both state-of-the-art text-based and multimodal KV cache methods including H2O~\cite{zhang2024h2o}, SnapKV~\cite{Li2024SnapKVLK}, PyramidKV~\cite{zhang2024pyramidkv}, LOOK-M~\cite{wan2024look}, and is able to achieve up to \textbf{2.82} times faster inference speed and reduce KV cache memory footprint by up to \textbf{72\%}, while maintaining or improving performance on the target tasks. 

\section{Related Work}
\vspace{-2mm}

\noindent \textbf{Post-training KV Cache Compression.} 
Post-training KV cache compression methods~\cite{wan2023efficient, liu2024contemporary} fall into four categories: token-wise eviction, token-wise merging, static layer-wise reduction, and quantization. Token-wise eviction (e.g., StreamingLLM~\cite{Xiao2023EfficientSL}) retains key tokens for sequence generation, while $\text{H}{2}\text{O}$~\cite{zhang2024h2o}, SnapKV \cite{Li2024SnapKVLK},ParallelComp~\cite{xiong2025parallelcomp}, and UNComp~\cite{xiong2024uncomp} focus on compact subsets, potentially sacrificing context. Token-wise merging (e.g., CaM~\cite{Zhang2024CaMCM}, $\text{D}{2} \text{O}$~\cite{wan2024d2o}) re-integrates tokens to maintain context. Static layer-wise reduction (e.g., PyramidKV~\cite{zhang2024pyramidkv}) linearly reduces cache across layers but ignores inter-layer attention variations. Quantization (e.g., KIVI~\cite{liu2024kivi}, Gear~\cite{kang2024gear}) balances memory and precision.
Most methods focus on text-based KV compression, overlooking multimodal contexts. LOOK-M~\cite{wan2024look} addresses multimodal compression but uses fixed allocation, neglecting inter-layer attention differences. MEDA introduces a multimodal attention entropy-guided dynamic allocation to address this.


\noindent \textbf{Vision Token Compression for MLLMs.} 
Classical approaches such as MobileVLM~\cite{chu2024mobilevlm}, LLaVA-Prumerge~\cite{Shang2024LLaVAPruMergeAT}, MADTP~\cite{Cao2024MADTPMA}, and FastV~\cite{Chen2024AnII} focus on reducing image tokens, which dominate the total token count, to accelerate inference by removing redundancies. MobileVLM~\cite{chu2024mobilevlm} uses a lightweight projector with average pooling to compress visual tokens, while LLaVA-Prumerge~\cite{Shang2024LLaVAPruMergeAT} and MADTP~\cite{Cao2024MADTPMA} adopt adaptive strategies to reduce tokens while preserving performance. FastV~\cite{Chen2024AnII} offers a plug-and-play solution that optimizes early layer computations and prunes visual tokens in later layers.
In contrast, MEDA focuses on multimodal KV cache compression through a dynamic layer-wise allocation strategy, eliminating the need for additional fine-tuning and enhance the efficiency of multimodal long-context generative inference.


\noindent \textbf{Long-context MLLMs.}  
Recent works have expanded MLLMs' multimodal long-context capabilities through additional training. \citet{liu2024world} leverage Blockwise RingAttention for scalable long-sequence training.  LongVA~\cite{zhang2024long} first pre-trains LLMs on long-text sequences and then aligns Long LLMs using short vision data to generalize to multimodal long-text contexts. LongLLaVA~\cite{wang2024longllava} modifies the model architecture by integrating Mamba and Transformer blocks and employs a progressive training strategy using multiple images. Video-XL~\cite{shu2024video} introduces visual context latent summarization to train models for handling even longer multimodal token sequences.
In contrast, MEDA introduces a dynamic KV cache optimization algorithm, enhancing long-context multimodal inference without additional training and is compatible with these methods.

\begin{figure*}[ht]
    \centering
    \includegraphics[width=0.9\textwidth]{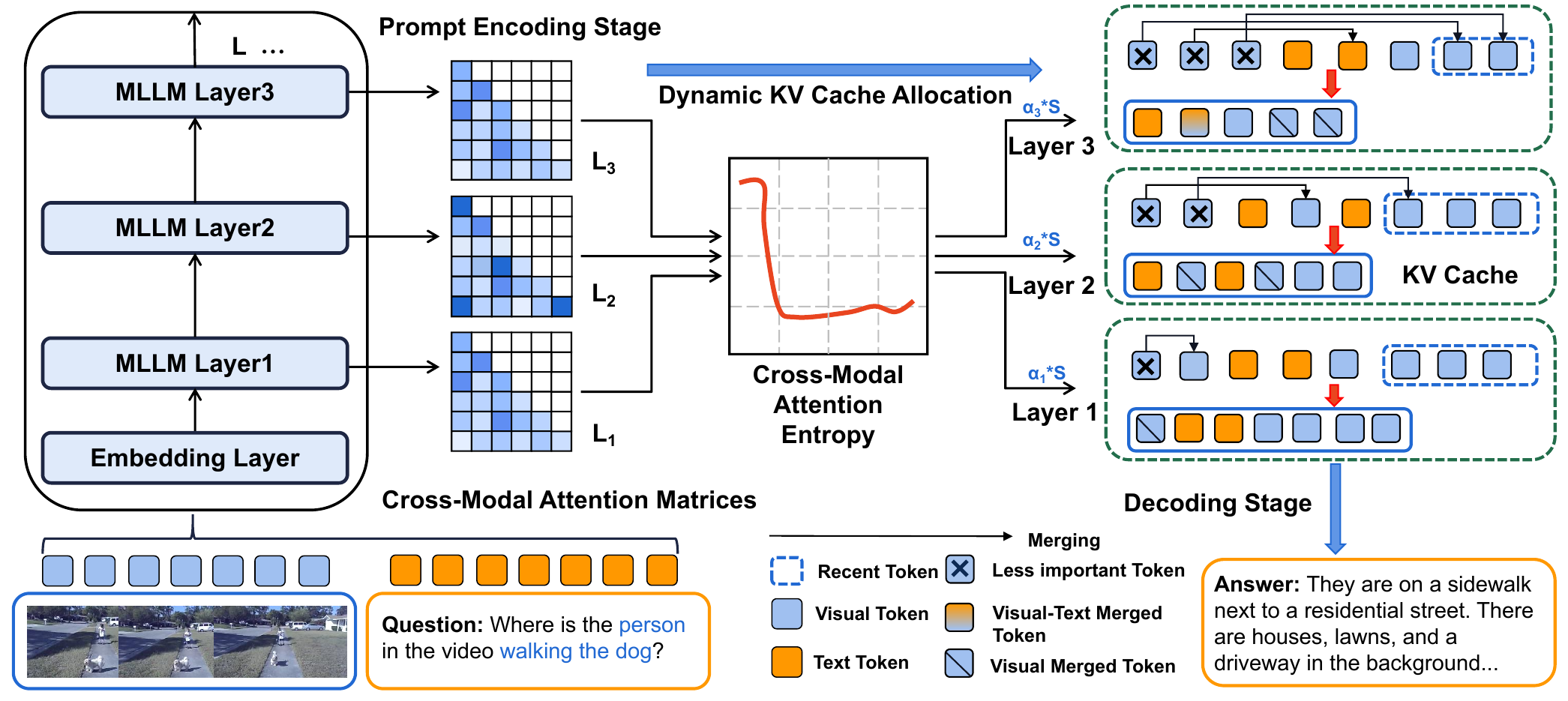}
    \caption{\small{ 
Illustration of MEDA's multimodal attention entropy-guided dynamic KV cache allocation and merging strategy.}}
    \label{fig: samplen}
\end{figure*}

\section{MEDA}


\subsection{Background on Generative Inference with Multimodal Context} \label{sec: Background}

Standard generative inference process of MLLMs involves two stages: (i) multimodal long-context prompt encoding, and (ii) decoding with multimodal KV cache.

\noindent \textbf{Multimodal Long-Context Prompt Encoding}. 
In the prompt encoding stage, a sequence of prompts comprising text, images, or videos is used to construct the Key-Value (KV) cache for each transformer layer in MLLMs. 
Specifically, let $\mathbf{X} \in \mathbb{R}^{L_{\text{prompt}} \times D}$ denote the input prompt tensor, where $L_{\text{prompt}}$ is the total length of the prompt sequence and $D$ is the hidden dimension of the model. The input prompt tensor can be expressed as:
$\mathbf{X} = \{\mathbf{X}^{T}_{1}, \mathbf{X}^{I}_{1}, \ldots, \mathbf{X}^{T}_{N}, \mathbf{X}^{I}_{M}\}$
or $\mathbf{X} = \{\mathbf{X}^{T}_{1}, \mathbf{X}^{T}_{2}, \ldots, \mathbf{X}^{V}_{p}, \mathbf{X}^{V}_{q}\}$
where $\mathbf{X}^{T}_{n}$, $\mathbf{X}^{I}_{m}$, and $\mathbf{X}^{V}_{q}$ represent the embeddings for the $n$-th text token, $m$-th image token, and $q$-th video token, respectively. 
In the text-multi-images setting, embeddings from different modalities are often interleaved in the input sequence. 
In the long-video setting, the number of video embeddings can become large due to the large number of input video frames, leading to a significant increase in decoding length.
For simplicity, we omit indices for attention heads and layers. The key and value tensors are computed as:
\begin{equation} 
\small
\mathbf{K} = \mathbf{X} \mathbf{W}_K, \quad \mathbf{V} = \mathbf{X} \mathbf{W}_V, 
\end{equation}
\noindent where $\mathbf{W}_K, \mathbf{W}_V \in \mathbb{R}^{D \times D}$ are the key and value projection matrices. The computed $\mathbf{K}$ and $\mathbf{V}$ are stored in the KV cache to facilitate subsequent token generation.

\noindent \textbf{Decoding with Multimodal KV Cache}. 
In the decoding stage, the KV cache is utilized and updated to generate tokens sequentially. At each time step $t$, the keys and values for the new token $\mathbf{x}_{t}$ are computed, while the keys and values for previous tokens $\mathbf{x}{<t}$ are retrieved from the cache. Denoting concatenation by $[\cdot]$, the KV cache is updated as:
\begin{equation} 
\small
\mathbf{K} = [\mathbf{K}, \mathbf{x}_{t} \mathbf{W}_K], \quad \mathbf{V} = [\mathbf{V}, \mathbf{x}_{t} \mathbf{W}_V].
\end{equation}
The output for the newly generated token is then computed as:
\begin{equation}
\small
    \mathbf{x}_{t, out} = \operatorname{Softmax}\left(\mathbf{q}_{t} \mathbf{K}^{\top} / \sqrt{D}\right) \mathbf{V}, \mathbf{q}_{t} = \mathbf{x}_{t} \mathbf{W}_Q,
\end{equation}
where $\mathbf{W}_Q \in \mathbb{R}^{D \times D}$ is the query projection. 

\noindent \textbf{Challenge.} The inclusion of multimodal long-context inputs and complex interactions between multimodal tokens (text, images, videos) significantly increases the size and complexity of the KV cache. Unlike text-only models, multimodal scenarios involve intricate cross-modal interactions between tokens, which pose new challenges for compressing the KV cache in long-context settings.

\subsection{Cross-Modal Attention Entropy for Dynamic KV Cache Allocation}
\label{sec: cross-modal entropy}
Cross-modal interactions in MLLMs create diverse attention patterns across MLLMs layers, and ignoring these variations leads to inefficient cache usage and degraded performance. Thus, designing a  multimodal dynamic KV cache allocation strategy that adapts to layer-wise attention distribution is crucial for efficient KV cache management. To capture the attention distribution characteristics across different layers, we introduce the concept of \textbf{cross-modal attention entropy}. Attention entropy~\cite{de2005tutorial, zhai2023stabilizing} quantifies the uncertainty or dispersion of the attention weights, providing insights into how focused or diffused the model's attention is at each layer.

As illustrated in Figure~\ref{fig: samplen}, for each layer $l$ in the MLLMs, we compute the \textbf{cross-modal attention matrices} between text and visual tokens.  Specifically, the attention from text to vision ($\mathbf{A}_{\text{TV}}^l$) and from vision to text ($\mathbf{A}_{\text{VT}}^l$) are calculated as:
\begin{equation}
\small
\begin{aligned} & \mathbf{A}_{\mathrm{TV}}^l=\operatorname{Softmax}\left(\mathbf{Q}_T^l\left(\mathbf{K}_V^l\right)^{\top} / \sqrt{D}\right), \\ & \mathbf{A}_{\mathrm{VT}}^l=\operatorname{Softmax}\left(\mathbf{Q}_V^l\left(\mathbf{K}_T^l\right)^{\top} / \sqrt{D}\right),
\end{aligned}    
\end{equation}
\noindent where $\mathbf{Q}_T^l \in \mathbb{R}^{n_T \times d}$ and $\mathbf{K}_T^l \in \mathbb{R}^{n_T \times D}$ are the query and key matrices for text tokens,
$\mathbf{Q}_V^l \in \mathbb{R}^{n_V \times D}$ and $\mathbf{K}_V^l \in \mathbb{R}^{n_V \times D}$ represent the query and key matrices for visual tokens, which are derived from the original $\mathbf{Q}$ and $\mathbf{K}$ based on the modality index, $n_{T}$ and $n_{V}$ are the numbers of text and visual tokens, and $D$ is the dimensionality of the embeddings.
We define the attention entropy of a row $i$ of $\mathbf{A}$ by $\operatorname{E}\left(\mathbf{A}_i\right)=-\sum_{j=1}^T \mathbf{A}_{[i, j]} \log \left(\mathbf{A}[i, j]\right)$. 
Let $\operatorname{E}(A)=\frac{1}{T} \sum_{i-1}^T \operatorname{Ent}\left(A_i\right)$ denote the attention entropy of $\mathbf{A}$, $T$ is the number of tokens and $\mathbf{A}$ is average attention across multi-heads for each layer. The cross-modal attention entropy 
$\mathbf{E}_{CM}^{l}$ for layer $l$ is then computed as:
\begin{equation}
\small
\begin{aligned} & \mathbf{E}_{TV}^{l} = \frac{1}{|T|}\sum_{i=1}^{n_T} \sum_{j=1}^{n_V} \mathbf{A}_{\mathrm{TV}}^l[i, j] \log \mathbf{A}_{\mathrm{TV}}^l[i, j], \\ & \mathbf{E}_{VT}^{l} = \frac{1}{|V|}\sum_{i=1}^{n_T} \sum_{j=1}^{n_V} \mathbf{A}_{\mathrm{VT}}^l[i, j] \log \mathbf{A}_{\mathrm{VT}}^l[i, j],
\end{aligned}
\end{equation}
\begin{equation}
\small
\mathbf{E}_{CM}^{l} = -(\mathbf{E}_{TV}^{l} + \mathbf{E}_{VT}^{l}),
\label{eq: Entropy}
\end{equation}
where $|T|$, $|V|$ denotes the number of text and visual tokens respectively. This entropy measures the uncertainty in the cross-modal attention distributions between text and visual tokens. A lower entropy indicates that the attention is more concentrated on specific cross-modal token pairs, suggesting that the layer is focusing on some more important multimodal interactions. 
Therefore, using the computed cross-modal attention entropy, we propose an inverse entropy softmax allocation strategy to determine the proportion $\alpha_{l}$ of the total KV cache size $S$  allocated to layer $l$: 
\begin{equation}
\small
S_l=\alpha_l \cdot S, \text{ }\alpha_l=\frac{\exp \left(\mathbf{E}_{\mathrm{CM}}^l\right)}{\sum_{k=1}^L \exp \left(\mathbf{E}_{\mathrm{CM}}^k\right)} \cdot L \cdot \rho, 
\end{equation}

\noindent where the attention entropy-guided dynamic allocated KV cache size for layer $l$ is $S_l$, $L$ is the total number of layers in the model, and $\rho \in(0,1]$ is the compression ratio representing the fraction of the original cache size to retain. The allocation strategy ensures that layers with \textit{lower} cross-modal attention entropy receive a \textit{smaller} portion of the KV cache, effectively preserving critical cross-modal information. Layers with \textit{higher} cross-modal attention entropy, indicating more diffused attention, receive \textit{larger} cache allocation. Such dynamic layer-wise KV cache allocation strategy optimizes memory usage.
%
Details of $\alpha_{l}$ are described in~\ref{sec: Details of Allocation Factor}.

\subsection{Multimodal KV Pair Selection and Merging}
\label{sec: Multimodal KV Cache Compression}
As illustrated in Figure~\ref{fig: samplen}, given the dynamically allocated KV cache size $S_l$ at layer $l$,  MEDA employs a multimodal KV pair selection scheme to identify which KV pairs to be selected at layer $l$.  
For each KV pair that is not selected, MEDA incorporates a KV pair merging strategy that merges the selected and non-selected KV pairs to preserve information from the entire context despite the reduced KV cache size.


\noindent \textbf{Multimodal KV Pair Selection.} 
Text tokens often contain crucial semantic information in multimodal contexts~\cite{wan2024look}.  Therefore, during the prompt encoding stage, we prioritize the retention of text-based KV pairs and select conserved tokens based on the compression ratio and accumulated attention ranks. The tokens not selected are considered as \textit{less important tokens}. Unlike previous eviction-based~\cite{zhang2024h2o} or static layer-wise reduction methods~\cite{zhang2024pyramidkv, yang2024pyramidinfer} that discard these less important tokens, we employ KV cache merging techniques to preserve the integrity of the contextual information. The cumulative attention score $\mathbf{A}_{s}$ is computed as:
\begin{equation}
\small
\mathbf{A}_s=\sum_{i=1}^{L_{\text {prompt }}} \mathbf{A}_p[i,:], \quad \mathbf{A}_p=\operatorname{Attn}\left(\mathbf{Q}_p \mathbf{K}_p^{\top}\right),
\end{equation}
where $\mathbf{A}_p$ denotes the attention weights during prompt encoding, and $\mathbf{Q}_p, \mathbf{K}_p \in \mathbb{R}^{L_{\text {prompt }} \times D}$ are the query and key matrices of the prompt tokens, respectively. To prioritize text tokens, we enhance their attention scores by adding a max value among all tokens:
\begin{equation}
\small
\mathbf{A}_s[T]=\mathbf{A}_s[T]+\max \left(\mathbf{A}_s\right),
\end{equation}
where $T$ denotes the indices of text tokens. This adjustment ensures that text-based KV pairs are more likely to be retained during the selection process. 
%
We also retain a recent context window of size $M$ to preserve immediate context, and then select the top $N$ important tokens with the highest attention scores from the remaining tokens. The conserved KV cache ($\mathbf{K}_{c}, \mathbf{V}_{c}$) is given by: 
\begin{equation}
\small
\begin{gathered}
\mathbf{K}_c=[\mathbf{K}[I,:] ; \mathbf{K}[-M:,:]], \quad \mathbf{V}_c=[\mathbf{V}[I,:] ; \mathbf{V}[-M:,:]] \\
I=\operatorname{Top}_N\left(\mathbf{A}_s[:-M]\right)
\end{gathered}
\end{equation}
where $\operatorname{Top}_N(\cdot)$ selects the indices of the top $N$ tokens based on $\mathbf{A}_{s}$, excluding the most recent $M$ tokens. The tokens not included in ($\mathbf{K}_{c}, \mathbf{V}_{c}$) constitute less important KV pairs ($\mathbf{K}_{less}, \mathbf{V}_{less}$).

\noindent \textbf{Multimodal KV Pair Merging.} 
To preserve the integrity of the contextual information, we employ a KV pair merging strategy that integrates the less important tokens into the conserved cache, rather than discarding them as in prior works \cite{zhang2024pyramidkv, Li2024SnapKVLK}. 
Specifically, we perform a many-to-one nearest-neighbor matching~\cite{wan2024look} between the less important KV pairs ($\mathbf{K}_{less}, \mathbf{V}_{less}$) and the conserved KV pairs ($\mathbf{K}_{c}, \mathbf{V}_{c}$).
We compute the similarity matrix $\mathbf{U} \in \mathbb{R}^{L_{\operatorname{les}} \times L_c}$ between the keys of less important and conserved tokens using cosine similarity:
\begin{equation}
\small
\mathbf{u}_{i, j}=\frac{\mathbf{k}_i^{\top} \mathbf{k}_j}{\left\|\mathbf{k}_i\right\|\left\|\mathbf{k}_j\right\|}, \quad i \in I_{\text {less }}, j \in I_c
\end{equation}
where $\mathbf{k}_i \in \mathbf{K}_{\text {less, }} \mathbf{k}_j \in \mathbf{K}_c$, and $I_{\text {less }}, I_c$ are the indices of less important and conserved tokens, respectively. For each less important token $\mathbf{k}_{i}$, we identify its nearest conserved token by finding the one with the highest similarity as $j=\arg \max _{j \in I_c} \mathbf{s}_{i, j}$. We then merge the less important tokens with their corresponding conserved tokens. In MEDA, we adopt the \textbf{average merging}  strategy, where we update the conserved key as:
\begin{equation}
\small
\mathbf{k}_j \leftarrow \frac{1}{\left|\mathcal{N}_j\right|+1}\left(\mathbf{k}_j+\sum_{i \in \mathcal{N}_j} \mathbf{k}_i\right).
\end{equation}
Considering the alignment properties of KV pairs~\cite{Zhang2021AlignmentAB}, we compute the similarity matrix only on the key tokens and apply the same similarity metric and the same averaged merging to the value tokens.
Therefore, by merging the less important tokens into the conserved ones, we aim to preserve essential information from the entire context, thus maintaining coherence during decoding despite the reduced cache size. 

\begin{table*}[t]
\centering
\caption{\small{Performance of various KV cache strategy on several MLLMs on MileBench's tasks with compression ratio $\rho=0.1$. A-Merge, W-Merge, P-Merge denote averaged merging, weighted merging and pivotal merging, respectively. TR represents text-prior KV pair eviction.}}
\label{tab: milebench}
\small
\scalebox{0.90}{
\begin{tabularx}{\textwidth}{l|*{11}{>{\centering\arraybackslash}X}}
\toprule
Method & T-1 & T-2 & T-3 & T-4 & S-1 & S-2 & S-3 & S-4 & S-5 & NH & IR \\ 
\midrule
\multicolumn{12}{c}{LLaVA-NeXT-7B} \\
\midrule
\rowcolor{gray!16}
\textbf{Full Cache} & 45.8 & 51.8 & 38.3 & 44.8 & 61.1 & 39.6 & 18.7 & 29.8 & 66.9 & 5.5 & 7.6 \\
\midrule
$\textbf{H2O}$~\cite{Zhang2023H2OHO} & 42.0 & 48.8 & 35.1 & 41.5 & 57.8 & 35.9 & 15.0 & 25.9 & 63.8 & 2.4 & 3.7 \\ 
\textbf{SnapKV}~\cite{Li2024SnapKVLK}& 42.1 & 48.5 & 35.2 & 41.1 & 57.6 & 36.1 & 15.6 & 26.6 & 63.3 & 1.8 & 3.2 \\ 
\textbf{PyramidKV}~\cite{zhang2024pyramidkv} & 42.4 & 48.6 & 35.1 & 41.7 & 57.9 & 36.5 & 15.2 & 26.7 & 63.6 & 2.0 & 3.2 \\
\textbf{LOOK-M}~\cite{wan2024look} & 44.1 & 50.1 & 36.7 & 43.0 & 59.5 & 37.8 & 17.2 & 28.1 & 65.3 & 3.6 & 5.0 \\
\rowcolor{blue!16}
\textbf{MEDA} (\textit{Ours}) & \textbf{45.4} & \textbf{50.9} & \textbf{37.6} & \textbf{44.1} & \textbf{60.2} & \textbf{38.9} & \textbf{17.8} & \textbf{30.1} & \textbf{66.9} & \textbf{4.8} & \textbf{7.4 }\\
\midrule
\multicolumn{12}{c}{InternVL-v1.5-7B} \\
\midrule
\rowcolor{gray!16}
\textbf{Full Cache}  & 10.7 & 19.2 & 13.8 & 19.1 & 16.8 & 9.7 & 14.4 & 19.1 & 5.0 & 11.1 & 0.0 \\
\midrule
$\textbf{H2O}$~\cite{Zhang2023H2OHO} & 8.4 & 16.8 & 11.5 & 16.7 & 14.6 & 7.3 & 12.1 & 16.8 & 2.7 & 8.9 & 0.0 \\
\textbf{SnapKV}~\cite{Li2024SnapKVLK} & 8.9 & 17.1 & 11.8 & 17.3 & 15.1 & 7.6 & 12.5 & 16.9 & 3.2 & 9.2 & 0.0 \\
\textbf{PyramidKV}~\cite{zhang2024pyramidkv} &8.3 & 17.7 & 12.0 & 16.8 & 15.1 & 8.2 & 11.9 & 16.8 & 3.1 & 9.3 & 0.0 \\
\textbf{LOOK-M}~\cite{wan2024look} & 9.2 & 17.5 & 12.2 & 17.8 & 15.1 & 8.5 & 12.9 & \textbf{19.2} & 4.0 & 10.2 & 0.0 \\
\midrule
\rowcolor{blue!16} \textbf{MEDA} (\textit{Ours}) & \textbf{10.5} & \textbf{18.8} & \textbf{13.3} & \textbf{18.6} & \textbf{16.4} & \textbf{9.0} & \textbf{14.3} & 18.4& \textbf{5.0} & \textbf{10.9} & \textbf{1.2} \\
\midrule
\multicolumn{12}{c}{LLaVA-v1.5-13B} \\
\midrule
\rowcolor{gray!16}
\textbf{Full Cache} & 39.8 & 46.2 & 30.8 & 48.1 & 64.8 & 48.5 & 13.6 & 28.4 & 60.0 & 12.0 & 1.0 \\ 
\midrule
$\textbf{H2O}$~\cite{Zhang2023H2OHO} & 37.4 & 43.5 & 28.2 & 45.6 & 61.4 & 45.0 & 11.3 & 25.5 & 57.6 & 9.8 & 0.0 \\ 
\textbf{SnapKV}~\cite{Li2024SnapKVLK} & 36.3 & 42.9 & 27.7 & 45.0 & 61.5 & 45.3 & 10.4 & 25.0 & 56.9 & 9.0 & 1.0 \\ 
\textbf{PyramidKV}~\cite{zhang2024pyramidkv} & 36.4 & 43.5 & 28.1 & 45.2 & 62.2 & 45.1 & 10.4 & 25.0 & 56.8 & 9.0 & 0.0 \\ 
\textbf{LOOK-M}~\cite{wan2024look} &38.8 & 44.3 & 28.5 & 46.5 & 62.8 & 46.4 & 11.7 & 26.7 & 57.6 & 10.2 & 1.0 \\
\midrule
\rowcolor{blue!16} \textbf{MEDA} (\textit{Ours}) & \textbf{40.0} & \textbf{46.2} & \textbf{30.4} & \textbf{48.0} & \textbf{64.7} & \textbf{48.6} & \textbf{13.1} & \textbf{27.6} & \textbf{59.7} & \textbf{12.2} & \textbf{1.5} \\

\bottomrule
\end{tabularx}
}
\vspace{-0.2in}
\end{table*}

\section{Experimental Setups}
\vspace{-1mm}

\subsection{Datasets and Evaluation Metrics}
To evaluate the performance of MEDA,
we conduct experiments on two types of datasets: text-multi-images datasets and long-video datasets. 

\noindent \textbf{For text-multi-images datasets}, we use the MileBench~\cite{Song2024MileBenchBM} benchmark, which includes four categories of tasks: Temporal Multi-image Tasks (T), Semantic Multi-image Tasks (S), Needle in a Haystack Tasks (NH), and Image Retrieval Tasks (IR). Performance is assessed using metrics including accuracy and ROUGE-L.

\noindent \textbf{For long-video datasets}, we use generative evaluation benchmarks Video-ChatGPT~\cite{maaz2023video} and DREAM-1K~\cite{wang2024tarsier} for video description, and generation-based open-ended QA dataset WorldQA~\cite{zhang2024worldqa}. The quality of generated responses is evaluated using the GPT API\footnote{https://platform.openai.com/docs/models}. Additional details regarding these datasets are provided in Appendix~\ref{sec: dataset desription}.

\vspace{0mm}
\subsection{Models}
\vspace{-1mm}

To evaluate MEDA's robustness across different model architectures, we conduct experiments on both multimodal long-context and long-video models.
For multimodal long-context models, we evaluate LLaVA-v1.5-13B~\cite{Liu2023VisualIT}, LLaVA-NeXT-7B~\cite{liu2024llava}, and InternVL-v1.5-7B~\cite{Chen2023InternVLSU}. 
For long-video models, we evaluate LLaVA-Video-7B/32B~\cite{zhang2024video}, LongVA-7B~\cite{zhang2024long}, and LongVILA-8B~\cite{xue2024longvila}.

\vspace{-1mm}
\subsection{Baselines} 
\vspace{-1mm}

To demonstrate the advantages of MEDA, we use the latest KV cache compression methods as baselines. \textbf{H2O}~\cite{zhang2024h2o} and \textbf{SnapKV}~\cite{Li2024SnapKVLK} employ eviction-based strategies, while \textbf{PyramidKV}~\cite{zhang2024pyramidkv} uses static layer-wise reduction. As these methods are text-centric, we adapt them for fair comparison in multimodal long-context settings. We also include \textbf{LOOK-M}~\cite{wan2024look}, the state-of-the-art multimodal KV cache compression method.

\vspace{-1mm}
\subsection{Implementation Details} 
\vspace{-1mm}

Let $\rho$ denote the total compression ratio of the KV cache across layers, with the compressed cache as $\rho \times \textit{input\_context}$. For each layer, the allocated KV cache is $\alpha_l \times \textit{input\_context}$, where $\alpha_l = \beta_1 + \beta_2$. We followed~\cite{wan2024look} to set the ratio $\beta_1 : \beta_2$ to $3:1$ that corresponds to recent context tokens $M$ and important tokens $N$, respectively, with the memory overhead per layer proportional to $\beta_1 + \beta_2$.
All of the experiments are conducted on NVIDIA A100 GPUs.

\section{Experiment Results}
\label{sec: exp_result}
\vspace{-2mm}

We present experiments demonstrating MEDA's effectiveness in optimizing multimodal KV cache across various models. Evaluations on MileBench~\cite{Song2024MileBenchBM} highlight its benefits in long-context scenarios, while tests on long-video datasets confirm its generative reasoning capabilities. We analyze the impact of KV cache budgets at different compression ratios $\rho$ on token generation and provide ablation studies for further insight into MEDA's performance. Additionally, we evaluate computational efficiency by measuring runtime and KV cache load during decoding.

\subsection{Performance on MileBench}
\label{sec: milebench}
To validate the effectiveness of MEDA in dynamically compressing KV cache in multi-image and multi-text cross-attention scenarios, we conducted experiments on MileBench using various models ranging from 7B to 13B parameters. We set the compression ratio to $\rho = 0.1$ to compare the performance of different algorithms under low compression settings. As shown in Table~\ref{tab: milebench}, MEDA outperforms other text-centric baselines across all models, such as H2O and SnapKV, demonstrating that incorporating attention differences between text and vision modalities helps preserve critical information during compression.
Moreover, MEDA surpasses PyramidKV and LOOK-M, indicating that it provides superior layer-wise cache allocation by dynamically adjusting the KV cache budget based on attention entropy between modalities, compared to uniform or progressive reduction techniques. The use of the KV Pairs Average Merging strategy allows MEDA to retain essential multi-modal tokens while eliminating redundant context.

\begin{table}[t]
\centering
\vspace{-0.1in}
\caption{\small{Performance of various KV cache strategy on Video-ChatGPT tasks with compression ratio $\rho=0.2$.} }
\label{tab: video_chatgpt}
\small
\scalebox{0.85}{
\begin{tabularx}{\linewidth}{l|*{4}{>{\centering\arraybackslash}X}}
\toprule
Method & Correct($\uparrow$) & Detail($\uparrow$)  & Context($\uparrow$) & Temp($\uparrow$) \\ 
\midrule
\multicolumn{5}{c}{LongVILA-8B} \\
\midrule
\rowcolor{gray!16}
\textbf{Full Cache} & 2.35 & 2.43 & 2.82 & 2.12 \\
\midrule
$\textbf{H}_{\textbf{2}}\textbf{O}$ & 1.96 & 2.08 & 2.38 & 1.66 \\ 
\textbf{SnapKV}& 1.84 & 2.12 & 2.38 & 1.72  \\ 
\textbf{PyramidKV} & 2.13 & 2.12 & 2.43 & 1.84  \\
\textbf{LOOK-M} & 2.19 & 2.15 & 2.51 & 1.88  \\
\rowcolor{blue!16}
\textbf{MEDA} (\textit{Ours}) & \textbf{2.25} & \textbf{2.29} & \textbf{2.69} & \textbf{2.05} \\
\midrule
\multicolumn{5}{c}{LongVA-7B} \\
\midrule
\rowcolor{gray!16}
\textbf{Full Cache} & 2.24 & 2.48 & 2.68 & 2.09 \\
\midrule
$\textbf{H2O}$ & 1.93 & 2.23 & 2.23 & 1.74 \\ 
\textbf{SnapKV}& 1.86 & 2.13 & 2.43 & 1.70  \\ 
\textbf{PyramidKV} & 1.91 & 2.08 & 2.35 & 1.78  \\
\textbf{LOOK-M} & 2.09 & 2.21 & 2.44 & 1.88 \\
\rowcolor{blue!16}
\textbf{MEDA} (\textit{Ours}) & \textbf{2.16} & \textbf{2.41} & \textbf{2.54} & \textbf{1.98} \\

\bottomrule
\end{tabularx}
}
\vspace{-0.2in}
\end{table}

\subsection{Performance on Long Video Tasks}
Next, we evaluate the generality and effectiveness of MEDA on various long-video MLLMs as well as video captioning and generative QA datasets, with model sizes ranging from 7B to 32B.


\noindent \textbf{Results on Video-ChatGPT. }
We evaluate each MLLM's predictions for correctness, detail orientation, and contextual and temporal understanding with a compressed KV cache at $\rho=0.2$.
Following prior settings~\cite{maaz2023video, xue2024longvila}, we use GPT-3.5 with 32 video frames to test each compression algorithm’s video captioning performance against human-annotated ground truth. As shown in Table~\ref{tab: video_chatgpt}, MEDA consistently outperforms text-centric baselines like H2O, SnapKV, and PyramidKV on LongVILA-8B and LongVA-7B models, showing that preserving key tokens from cross-modal interactions is crucial for long-video compression. MEDA also surpasses LOOK-M in all evaluation dimensions, confirming its effectiveness in dynamically adjusting KV cache size based on cross-modal attention entropy to retain essential multimodal information.


\begin{table}[t]
\centering
\caption{\small{Performance of various KV cache strategy on DREAM-1K tasks with compression ratio $\rho=0.2$.} }
\label{tab: dream1k}
\small
\scalebox{0.85}{
\begin{tabularx}{\linewidth}{l|*{3}{>{\centering\arraybackslash}X}}
\toprule
Method & F1($\uparrow$) & Precision($\uparrow$)  & Recall($\uparrow$)  \\ 
\midrule
\multicolumn{4}{c}{LLaVA-Video-7B} \\
\midrule
\rowcolor{gray!16}
\textbf{Full Cache} & 32.5 & 37.9 & 28.4  \\
\midrule
$\textbf{H2O}$ & 27.7 & 33.9 & 23.4  \\ 
\textbf{SnapKV}& 28.8 & 34.7 & 24.6   \\ 
\textbf{PyramidKV} & 28.1 & 34.2 & 23.9  \\
\textbf{LOOK-M} & 30.4 & 35.8 & 26.4  \\
\rowcolor{blue!16}
\textbf{MEDA} (\textit{Ours}) & \textbf{31.3} & \textbf{36.8} & \textbf{27.2}   \\
\midrule
\multicolumn{4}{c}{LLaVA-Video-32B} \\
\midrule
\rowcolor{gray!16}
\textbf{Full Cache} & 32.9 & 34.4 & 31.6	  \\
\midrule
$\textbf{H2O}$ & 28.1 & 28.8 & 27.5  \\ 
\textbf{SnapKV}& 29.1 & 30.6 & 27.7    \\ 
\textbf{PyramidKV} & 29.7 & 31.4 & 28.1   \\
\textbf{LOOK-M} & 29.6& 31.1 & 28.3 \\
\rowcolor{blue!16}
\textbf{MEDA} (\textit{Ours}) & \textbf{31.7} & \textbf{33.3} & \textbf{30.3}  \\

\bottomrule
\end{tabularx}
}
\end{table}

\begin{table}[t]
\centering
\vspace{0mm}
\caption{\small{Performance of different KV cache strategies on WorldQA tasks with a compression ratio of $\rho=0.2$. For instance, F represents Full Cache, and M represents MEDA.}  }
\label{tab: worldqa}
\small
\scalebox{0.85}{
\begin{tabularx}{\linewidth}{l|*{6}{>{\centering\arraybackslash}X}}
\toprule
\multicolumn{7}{c}{Open-ended QA($\uparrow$) } \\
\midrule
Models & \cellcolor{gray!16}\textbf{F} & \textbf{H} & \textbf{S} & \textbf{P} & \textbf{L} & \cellcolor{blue!16}\textbf{M}  \\ 
\midrule
LLaVA-Video-7B & \cellcolor{gray!16}27.3 & 24.3 & 24.5 & 23.8 &  24.9 &\cellcolor{blue!16} \textbf{25.8} \\
LongVA-7B & \cellcolor{gray!16}28.9 & 26.5 & 26.6 & 25.4 & 27.0 & \cellcolor{blue!16}\textbf{27.5}\\
LongVILA-8B & \cellcolor{gray!16}29.1 & 25.5 & 24.4 & 25.6 &  26.4 &\cellcolor{blue!16} \textbf{27.2}\\
\bottomrule
\end{tabularx}
}
\vspace{-0.2in}
\end{table}

\vspace{1mm}
\noindent \textbf{Results on DREAM-1K.} 
We evaluate the DREAM-1K~\cite{wang2024tarsier} benchmark on LLaVA-Video-7B and 32B models, using AutoDQ to assess five types of video data and report the average score. As shown in Table~\ref{tab: dream1k}, MEDA outperforms other KV cache compression strategies, maintaining performance close to that of the Full Cache even when compressed to 20\% on both 7B and 32B models. This result demonstrates that combining cross-modal attention entropy for layer-wise KV cache allocation with KV merging effectively preserves critical information in the KV cache.


\noindent \textbf{Results on WorldQA.} 
We further evaluate MEDA's reasoning capabilities on a generation-based open-ended QA dataset~\cite{zhang2024worldqa}, using GPT-4 to assess the quality of generated responses. As shown in Table~\ref{tab: worldqa}, MEDA outperforms other KV cache compression strategies across various model architectures, especially demonstrating a significant advantage over the eviction-based H2O method. The results confirm that MEDA’s combined dynamic layer allocation and average merging strategies can effectively maintain high-quality video reasoning even under high compression rates.

\begin{figure}[h]
\vspace{-0.1in}
\centering
\includegraphics[width=\linewidth]{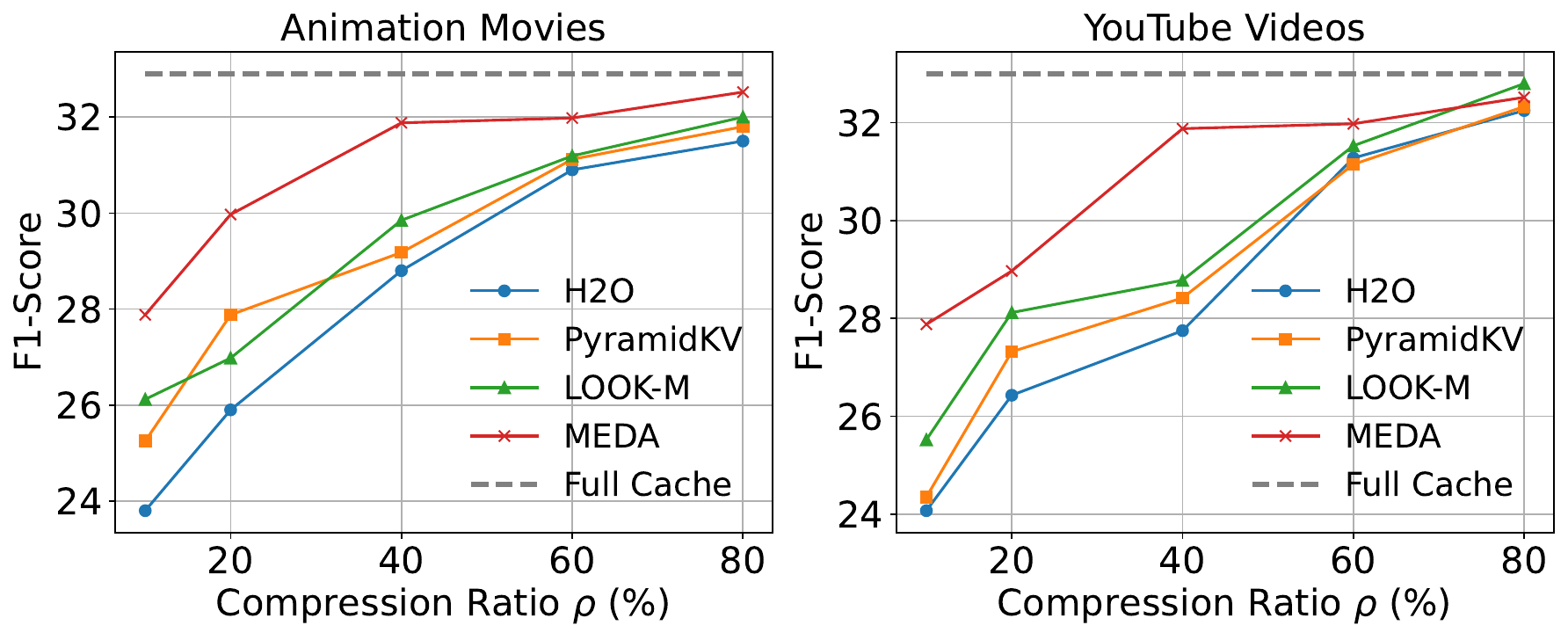}
\caption{Impact of compression ratio $\rho$.}
\vspace{0mm}
\label{fig:cachebudget}
\end{figure}

\subsection{Impact of Compression Ratio $\rho$}
We evaluate MEDA's efficiency across KV cache compression ratios $\rho$ from 10\% to 80\%, using standardized tests on the LLaVA-Video-7B model for video captioning tasks (live-action, animation, YouTube, and TikTok) from the DREAM-1K dataset.
As shown in Figure~\ref{fig:cachebudget}, MEDA consistently outperforms all baseline methods across all compression ratios $\rho$. Notably, at a compression ratio of 10\%, MEDA significantly surpasses both LOOK-M and PyramidKV, demonstrating its ability to dynamically preserve crucial multimodal information even under highly compressed conditions, thereby minimizing the loss of video context.

\begin{table}[t]
\centering

\vspace{-0.2in}
\caption{\small 
Results of ablation studies. For CL-CH and SP-DI, we use ROUGE-L and $\rho=0.1$. For AM and YT, we use F1 and $\rho=0.2$.}
\vspace{-0.25cm}
\scalebox{0.67}{
\begin{tabular}{lcccc}
\toprule
 & \textbf{CL-CH} & \textbf{SP-DI} & \textbf{AM} & \textbf{YT} \\
\midrule
& \multicolumn{2}{c}{LLaVA-NeXT-7B} & \multicolumn{2}{c}{LLaVA-Video-7B} \\
\midrule
\rowcolor{gray!16}
\textbf{Full Cache} & 19.5 & 18.7 & 27.6 & 33.0 \\
\textbf{H2O} & 16.6 & 16.2 & 23.8 & 27.1 \\
\midrule
\rowcolor{blue!16}
MEDA & \textbf{18.9} & \textbf{18.2} & \textbf{26.7} & \textbf{31.8} \\
\textit{w.o. Dynamic Allocation} & 17.8 & 17.5 & 25.5 & 28.2 \\
\textit{w.o. Average Merging} & 18.2 & 17.3 & 25.8 & 29.3 \\
\textit{w.o. Both} & 17.0 & 16.5 & 24.5 & 27.2 \\
\bottomrule
\end{tabular}
}
\label{tab: ablation}
\vspace{0mm}
\end{table}
\vspace{-0.1in}

\subsection{Ablation Study}

To demonstrate the effectiveness and generality of each component of MEDA, we present a table comparing Full Cache, H2O, and the key components of MEDA, evaluated on generative subtasks from MileBench (CLEVR-Change (CL-CH), Spot-the-Diff (SP-DI)) and DREAM-1K (Animation (AM), YouTube (YT)). As shown in Table~\ref{tab: ablation}, the removal of any MEDA component leads to performance degradation. Even with two components removed, MEDA still slightly outperforms H2O, which relies purely on eviction. These results confirm that each MEDA component mitigates information loss caused by entropy-based dynamic layer allocation and multimodal average merging.

\begin{table}[t]
\centering
\vspace{-0.2in}
\captionsetup{justification=centering}
\caption{Latency and KV cache GPU memory usage.}
\vspace{-2mm}
\scalebox{0.75}{
\begin{tabular}{lccc}
\toprule
Method & Budget $\rho$ & Decoding Latency & GPU Memory \\ 
\midrule
\rowcolor{gray!16}
Full Cache & 100\% &  14.61 ms/token & 2.42 GiB \\ 
MEDA  & 50\% & 11.58 ms/token & 1.34 GiB\\
MEDA  & 35\% & 9.64 ms/token & 1.05 GiB \\
MEDA  & 20\% &  8.23 ms/token & 0.67 GiB \\
MEDA  & 5\% & \textbf{5.18 ms/token} & \textbf{0.37 GiB} \\
\bottomrule
\end{tabular}
}
\label{tab:model_performance}
\vspace{-0.2in}
\end{table}

\vspace{-0.1in}
\subsection{Decoding Efficiency Analysis}
\label{sec: decoding latency}

We evaluate MEDA's efficiency in Table~\ref{tab:model_performance}, comparing decoding speed and memory usage against the Full Cache using LLaVA-Video-7B~\cite{zhang2024video}. Tests on 20 randomly selected DREAM-1K video, using a single A100 GPU, show that our compressed model has significantly lower decoding latency, particularly for long text generation. Speed and memory usage were further analyzed from 50\% to 5\% KV cache budgets. Results show that GPU memory usage scales proportionally to the cache budget, for instance, with a 20\% budget reducing memory usage by 72\% compared to Full Cache. Higher compression ratios also reduce latency, demonstrating the efficiency of our method. More details about specific environment for speed and memory test are illustrated in Appendix~\ref{sec: speed test}.

\vspace{-0.1in}
\section{Conclusion}
\vspace{-0.1in}
In this paper, we introduced MEDA, an approach for dynamically allocating KV cache in long-context MLLMs. By leveraging attention entropy, MEDA efficiently adapts to the complexities of cross-modal interactions, dynamically allocating layer-wise cache sizes and preserving critical information using an average merging technique, while significantly reducing memory usage and improving decoding speed. Our experiments show that MEDA achieves substantial memory savings and faster inference without sacrificing performance, setting a new standard for efficient multimodal processing in tasks involving long text-image and text-video inputs. These results affirm the effectiveness of MEDA in addressing the computational challenges inherent to multimodal models.

\vspace{-0.1in}
\section{Limitation}
\vspace{-0.1in}
The limitations of our study arise from optimizing the KV cache of standard MLLMs without incorporating advanced LLMs techniques such as quantization, pruning, distillation, retrieval augmented generation or efficient attention mechanisms~\cite{lin2024rotation, gu2024minillm, Dao2022FlashAttentionFA, tao2024scaling, wang2024svd, shen2024famba, li2024uncertaintyrag, wang2024iot, xin2025v}, and for mutltimodal generation tasks~\cite{shen2025efficient, xiong2024autoregressive}, math reasoning~\cite{zhang2024llama, xiong2022self}. In future work, we plan to combine these techniques and applications with MEDA to achieve the highest possible degree of KV cache compression.

\vspace{-0.1in}
\section{Acknowledgment}
\vspace{-0.1in}
This study is supported in part by NSF Award NeTS-2312675.

\bibliography{main}

\appendix

\section{Appendix}
\label{sec:appendix}

\subsection{Details of Datasets}
\label{sec: dataset desription}
\subsubsection{Details of MileBench}
The MileBench dataset, introduced by~\citep{Song2024MileBenchBM}, is the first comprehensive benchmark specifically designed to evaluate the Multimodal Long-context capabilities of Multimodal Large Language Models (MLLMs)~\cite{huang2024evolver, zhu2024dglf, gong2024neuroclips, liu2024benchmarking, wan2024meit}, which is different from static classification tasks~\cite{wan2023text, chen2025recent}. It contains 6,440 multimodal long-text samples sourced from 21 datasets, averaging 15.2 images and 422.3 words per sample. The benchmark is divided into two main components: Realistic Evaluation and Diagnostic Evaluation.

\noindent \textbf{Realistic Evaluation} This component evaluates the model's ability to handle long multimodal contexts, focusing on tasks that involve understanding and reasoning over extended sequences of multimodal data. Key tasks include:
\begin{enumerate}
    \item \textbf{Action Understanding and Prediction:} Models are assessed on tasks like action localization, prediction, and sequence understanding, using accuracy as the performance metric.
    \item \textbf{Object and Scene Understanding:} Tasks involve recognizing object existence, interaction, moving attributes, and shuffling of objects across time and scenes, with accuracy as the primary metric.
    \item \textbf{Visual Navigation and Spatial Localization:} These tasks test a model’s ability to navigate environments and track moving directions.
    \item \textbf{Counterfactual Reasoning and State Change:} The model is challenged with tasks like inferring state changes, understanding character order, and handling scene transitions.
    \item \textbf{Knowledge-Grounded Question Answering (QA):} This category includes QA tasks based on webpages, textbooks, multimodal content, and long texts paired with images, all evaluated using accuracy.
    \item \textbf{Text-Rich Image QA:} Models answer questions based on text-rich images such as slides and OCR content.
    \item \textbf{Visual Relation Inference:} Tasks include visual change captioning and expressing visual relationships, measured using the ROUGE-L metric.
    \item \textbf{Dialogue:} Multimodal dialogues and conversational embodied dialogues are tested, with ROUGE-L and accuracy metrics used to evaluate performance.
    \item \textbf{Space Understanding:} The model's ability to comprehend spatial relationships is also tested, with accuracy as the metric.
\end{enumerate}
\noindent \textbf{Diagnostic Evaluation:} This component focuses on specific challenges related to retrieving key information from complex or noisy multimodal data. Tasks include:
\begin{enumerate}
    \item \textbf{Text Needle in a Haystack:} A task that requires the model to find specific information in large text corpora, using accuracy as the metric.
    \item \textbf{Image Needle in a Haystack:} Similar to the text-based task, but involving visual search in large datasets of images.
    \item \textbf{Image Retrieval:} The model is evaluated on its ability to retrieve relevant images from a set, based on accuracy.
\end{enumerate} 

\subsubsection{Details of Video-ChatGPT}

Video-ChatGPT~\cite{maaz2023video} introduce a video-based text Generation performance Benchmark specifically designed to assess the text generation capabilities of video-based conversation models. To achieve this, they compile a test set from the ActivityNet-200 dataset~\cite{Heilbron2015ActivityNetAL}, which contains videos with dense descriptive captions and human-annotated question-answer pairs. They also develop an evaluation pipeline using the GPT-3.5 model, which rates the models' generated predictions on a scale from 1 to 5 across five evaluation criteria:
\begin{enumerate}
    \item \textbf{Information Accuracy:} This criterion checks that the generated text accurately reflects the content of the video, without misinterpretation or misinformation.
    \item \textbf{Detail Orientation:} The metric assess the thoroughness and specificity of the model’s responses, ensuring they capture the essential details of the video rather than providing overly generic answers.
    \item \textbf{Contextual Understanding:}  This aspect evaluates the model’s ability to generate responses that are coherent with the overall context of the video. 
    \item \textbf{Temporal Comprehension:} The model's understanding of the chronological order of events within the video is examined, especially when responding to questions about sequences of events.
    \item \textbf{Consistency:} The evaluation also considers the consistency of the model's responses across different, but similar, questions or across various sections of the same video.
\end{enumerate}

\subsubsection{Details of DREAM-1K}

DREAM-1K~\cite{wang2024tarsier} is introduced as a benchmark specifically designed to challenge video understanding models through detailed video description tasks, focusing on Rich events, Actions, and Motions. Unlike conventional video captioning datasets, which typically offer only brief, single-sentence captions for each video, DREAM-1K addresses the need for more comprehensive evaluation by providing in-depth descriptions. It includes 1,000 video clips of varying complexities, drawn from five distinct sources: live-action movies, animated films, stock footage, YouTube videos, and TikTok-style short clips. Each clip is annotated with detailed descriptions by expert annotators.
To more effectively assess the quality of these detailed video descriptions, DREAM-1K employs the \textbf{AutoDQ (Automatic Description Quality)} metric instead of traditional metrics like CIDEr, which are less suited for evaluating rich, detailed captions. AutoDQ has demonstrated both its discriminative power and strong alignment with human evaluations, making it a more interpretable and reliable method for automatic assessment in this context

\subsubsection{Details of WorldQA}
WorldQA~\cite{zhang2024worldqa} introduces an Open-Ended Question Answering (QA) task to evaluate the video-based generative reasoning capabilities of multimodal language models. For the evaluation of open-ended question answering (QA), WorldQA 
 employs \textbf{GPT-4}  with a specific scoring system that compares the model's answer $A$ with the ground truth $G$. This system assigns a score based on the following categories:
\begin{itemize}
    \item (1) $A = G$: Correct answer, earning 1 point.
    \item (2) $A \cap G = \emptyset$: Incorrect answer, scoring 0 points.
    \item (3) $\emptyset \subset A \cap G \subset A \cup G$: Partially correct answer, receiving 0.3 points.
    \item (4) $A \subset G$, $A \neq G$: Incomplete but correct answer, earning 0.5 points.
    \item (5) $G \subset A$, $A \neq G$: Redundant answer, also receiving 0.5 points.
\end{itemize}
This structured scoring approach evaluates the model's responses based on correctness, completeness, and redundancy.

\subsection{Details of Speed and Memory Testing}
\label{sec: speed test}

To ensure reproducibility of the results presented in Section~\ref{sec: decoding latency}, we detail the hardware setup, sample selection, as well as the methods used for calculating speed and GPU memory consumption below:

\noindent \textbf{\textit{Hardware Specifications}}: The tests were conducted using an AMD EPYC 7643 48-Core Processor and an NVIDIA A100 GPU.

\noindent \textbf{\textit{Speed Test Sample Selection}}: The first 20 samples from the DREAM-1K~\cite{wang2024tarsier} YouTube videos  dataset were selected for inference.

\noindent \textbf{\textit{Speed Measurement Methodology}}: The measured speed refers to the model’s decoding rate, beginning from the start of the decoding process. It is calculated by dividing the time spent in the decoding phase by the total number of tokens generated.

\noindent \textbf{\textit{Memory Usage Calculation Methodology}}: The average memory usage refers to the average memory allocated to the KVCache. The formula for this is given by:

\begin{equation*}
\begin{aligned}
\text{Memory Usage} = & \left( \textit{input\_len} + \textit{decoding\_len} \right) \\
& \times 2 \times 32 \times 32 \times \frac{128 \times 2}{1024^3} \, \text{GiB}
\end{aligned}
\end{equation*}

\noindent where input\_len represents the sequence length after pruning during the pre-filling phase, and decoding\_len represents the total number of tokens generated. The factor of 2 refers to the 2 bytes used for FP16 precision, 32 corresponds to both the number of attention heads and the number of layers in the model, and 128 is the dimensionality of each attention head. The second factor of 2 accounts for the storage of both Key and Value tokens.

\subsection{Details of Allocation Factor \(\alpha_l\) Derivation }
\label{sec: Details of Allocation Factor}
To derive the formula for the entropy-based softmax allocation strategy, we determine the proportion \(\alpha_l\) of the total KV cache size \(S\) allocated to each layer \(l\) based on the computed cross-modal attention entropy \(\mathbf{E}_{\text{CM}}^l\). The objective is to allocate more resources to layers that exhibit higher importance according to their entropy values, thereby preserving critical information.
Let \( S \) be the total KV cache size available across all layers, and \( L \) be the total number of layers in the model. We want to distribute this total cache size dynamically based on the significance of each layer, as measured by the cross-modal attention entropy \(\mathbf{E}_{\text{CM}}^l\). We denote the proportion of the total cache allocated to layer \( l \) as \(\alpha_l\), and the allocated cache size for that layer as \( S_l \). Therefore:
\begin{equation}
    S_l = \alpha_l \cdot S
\end{equation}
The cross-modal attention entropy \(\mathbf{E}_{\text{CM}}^l\) measures the concentration or dispersion of attention at layer \( l \). A higher entropy suggests that the layer has a more distributed attention pattern, indicating the need for more resources to capture and store these interactions. Thus, we define \(\alpha_l\) using a softmax function over the entropy:
\begin{equation}
    \alpha_l = \frac{\exp(\mathbf{E}_{\text{CM}}^l)}{\sum_{k=1}^L \exp(\mathbf{E}_{\text{CM}}^k)}
\end{equation}
This ensures that layers with higher entropy receive a greater proportion of the total cache. However, this normalization alone only provides weights that sum to 1 across all layers.

The factor \( L \), representing the number of layers, scales the normalized weights to ensure that the total allocation reflects the model's structure. The compression ratio \( \rho \in (0, 1] \) indicates the fraction of the original total cache size to be retained after compression, ensuring that the overall allocation respects the intended compression level.
Thus, the adjusted formula for \(\alpha_l\) becomes:
\begin{equation}
\alpha_l = \frac{\exp(\mathbf{E}_{\text{CM}}^l)}{\sum_{k=1}^L \exp(\mathbf{E}_{\text{CM}}^k)} \cdot L \cdot \rho    
\end{equation}
Finally, the cache size for each layer \( l \) is:
\begin{equation}
S_l = \alpha_l \cdot S = \left( \frac{\exp(\mathbf{E}_{\text{CM}}^l)}{\sum_{k=1}^L \exp(\mathbf{E}_{\text{CM}}^k)} \cdot L \cdot \rho \right) \cdot S    
\end{equation}
This formula dynamically distributes the total cache \( S \) across all layers, allocating more cache to layers with higher entropy, thus ensuring that critical cross-modal interactions are effectively captured. The strategy respects the general cache size and compression constraints while adapting based on the attention distribution of each layer, optimizing the allocation for efficient long-context inference in multimodal models.

\subsection{Generation Sample}

\begin{figure}[t]
    \centering
    \includegraphics[width=0.475\textwidth]{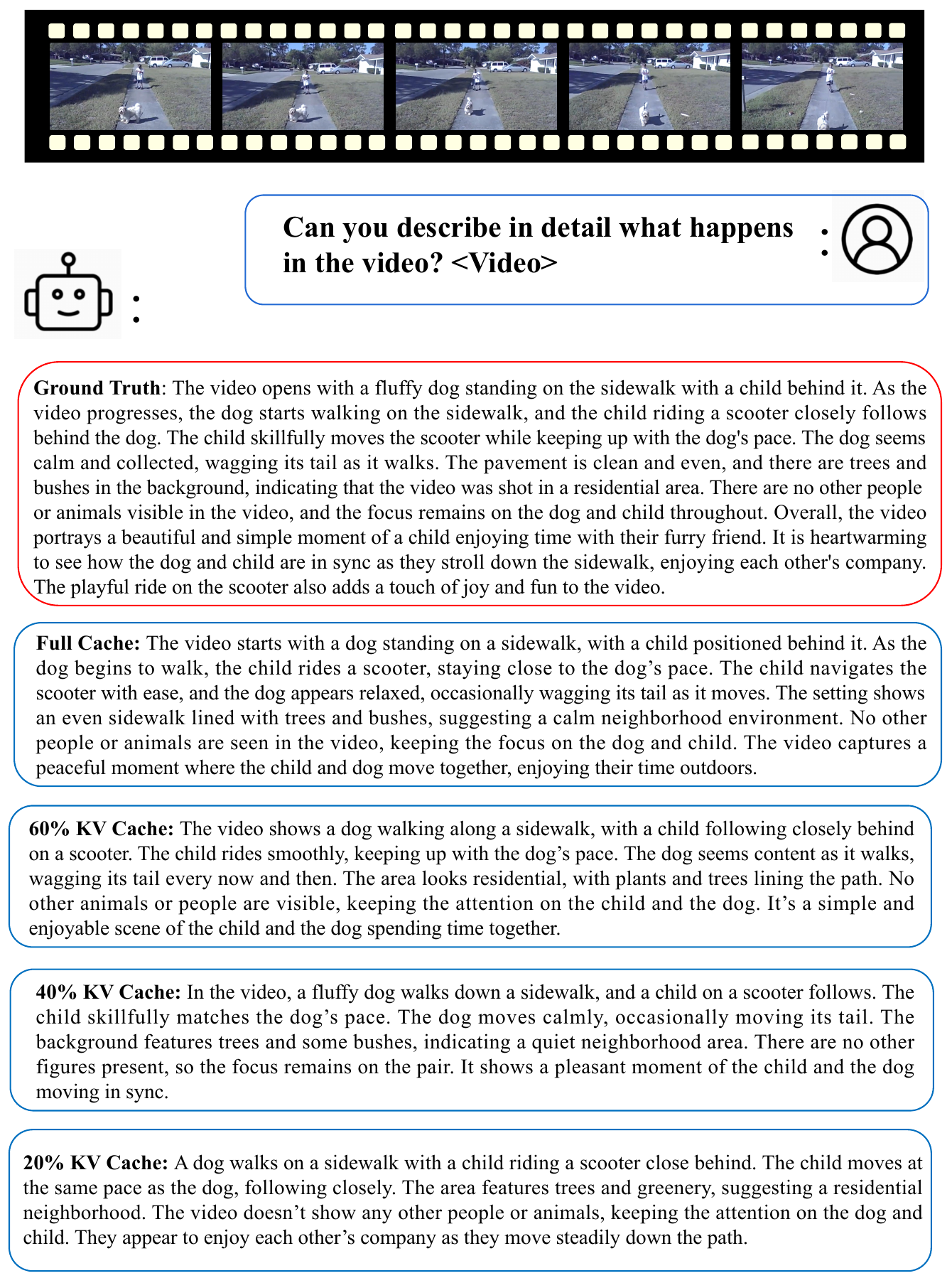}
   
    \caption{\small{An example of video content understanding and question answering based on Video-ChatGPT using the LongVA model and the KV cache compression technique of MEDA.}}

    \label{fig: sampel dec}
\end{figure}

To demonstrate the effectiveness of MEDA in generating responses under varying compression rates in video scenarios, we utilize the LongVA~\cite{zhang2024long} model and sample a result from Video-ChaGPT~\cite{maaz2023video}. As illustrated in Figure~\ref{fig: sampel dec}, we present the outputs of the LongVA model, ranging from the Full cache setting to a 20\% compression using MEDA. Our findings show that, with multimodal KV cache compression at levels of 60\% and 20\%, MEDA maintains a strong understanding of the video content and provides accurate responses. Remarkably, even with a 20\% compression of video or text tokens, the model employing MEDA produces outputs that are nearly semantically identical to those from the Full Cache configuration, also demonstrating an understanding of the video content that is close to the Ground Truth.

\end{document}